# A FUZZY BASED APPROACH TO TEXT MINING AND DOCUMENT CLUSTERING

Sumit Goswami[1] and Mayank Singh Shishodia[2]

[1]Indian Institute of Technology-Kharagpur, Kharagpur, India

sumit_13@yahoo.com

[2]School of Computer Science and Engineering, VIT University, Vellore, India

mayanksshishodia@gmail.com

## ABSTRACT

*Fuzzy logic deals with degrees of truth. In this paper, we have shown how to apply fuzzy logic in text mining in order to perform document clustering. We took an example of document clustering where the documents had to be clustered into two categories. The method involved cleaning up the text and stemming of words. Then, we chose 'm' features which differ significantly in their word frequencies (WF), normalized by document length, between documents belonging to these two clusters. The documents to be clustered were represented as a collection of 'm' normalized WF values. Fuzzy c-means (FCM) algorithm was used to cluster these documents into two clusters. After the FCM execution finished, the documents in the two clusters were analysed for the values of their respective 'm' features. It was known that documents belonging to a document type 'X' tend to have higher WF values for some particular features. If the documents belonging to a cluster had higher WF values for those same features, then that cluster was said to represent 'X'. By fuzzy logic, we not only get the cluster name, but also the degree to which a document belongs to a cluster.*

## 1. INTRODUCTION

There is an increasing amount of textual data available every day. Text Mining is the automated process of extracting some previously unknown information from analysis of these unstructured texts [1]. Although Text Mining is similar to Data Mining, yet they are quite different. Data mining is related to pattern-extraction from databases while for text-mining, it is natural language. It is also different from search, web search in particular, where the information being searched already exists. In Text Mining, the computer is trying to extract something new from the already available resources.

In this review paper, we have applied fuzzy logic to text-mining to perform document clustering into a number of pre-specified clusters. As an example, we have shown how to classify given documents into two categories – sports and politics. First, the documents are cleaned by removal of advertisements and tags. The hyphens are dealt with and the stop words are removed. Word stemming is carried out to represent each word by its root [2]. Each document is then represented as a bag of words, a method which assigns weights to words on basis of their significance in the document which we considered to be their frequency of occurrence. We calculated word frequency as

WF = (Word Count/ (Total Words in the Document)) x10000     (1)

'm' number of words are chosen[3][4][5] and each document is represented as a set of 'm' values which are the WF values of those corresponding 'm' words in that document. We then perform FCM on this representation to cluster the documents into required number of categories. Pre-known knowledge is used to name these clusters. In the forthcoming sections, we explain what fuzzy logic is, how it differs from probability, the FCM algorithm and how it can be used to perform document clustering with an example. We tell why fuzzy logic should be used for this purpose.

## 2. FUZZY LOGIC

Fuzzy logic is a mathematical logic model in which truth can be partial i.e. it can have value between 0 and 1, that is completely false and completely true [6]. It is based on approximate reasoning instead of exact reasoning.

### 2.1 Fuzzy Logic vs Probability

Probability can be divided into two broad categories – Evidential and Physical. Consider an event E with probability 80%. Evidential Probability says that given the current available evidence, the likelihood of E occurring is 80%. On the other hand, Physical Probability says that if the experiment involving E as an outcome was repeated an infinite number of times, then 80% times, E will occur [7].

For example, consider a car 'A' that goes 150 Kilometres per hour. Now this car would definitely be considered fast. Now consider another car 'B' that goes 180 Kilometres per hour. In a repeatable experiment, it is not that the car 'B' will be fast more often than A [Physical Probability] or that given the current information; it is more likely that Car 'B' is faster than Car 'A' [Evidential Probability]. Both the cars possess the property fast, but in different degrees.

If we use Fuzzy Logic for this problem and there is a class called "Fast Cars" that contains the cars with the property "fast" , then the respective membership values of both Car A and Car B can be compared. The higher membership value means that the corresponding car possesses the property "fast" to a higher degree.

### 2.2 Degrees of Truth

The degrees of truth are obtained from the membership function. An example is shown in Figure1:

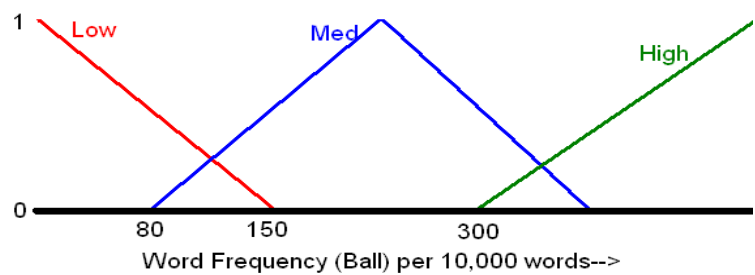

Figure 1: Membership functions for word frequency of "ball"

Suppose, we are given a statement that high frequency of occurrence of the word "Ball" in a document means that the document is related to Sports. In order to associate numerical values with these "high" or similarly "low", "medium" frequency of words, we use a membership function. Linguistic hedging can be used to create more categories by adding adjectives like "Very", "Fairly", "Not so", "Less", "More" etc.

Figure1 shows a membership function which decides when the frequency of occurrence of the word "ball" in a document is said to be low, medium or high. But it should be noted that the membership function depends on the situation and problem as perceived by the designer. For example, someone else may design the membership function where the word frequency of even 500 is considered low. This can be the case for a document containing the ball-by-ball commentary of a cricket match. A sample portion of this document could be like "*The captain has handed over the ball to Player A for the final over. He runs in to bowl the first ball. It's a short ball outside the off stump. It's a no-ball! The batsman smashes the ball and it goes for a sixer*". Here, the word ball occurs 5 times in total of 44 words. This is a 1136 word frequency (per 10000 words) value.

## 3. CLUSTERING

Using Fuzzy Logic and text-mining we can cluster similar documents together. Document Clustering is used by the computer to group documents into meaningful groups. We will use c-means algorithm. This algorithm is classified into two types- Hard c-means and FCM [6].

Hard c-means algorithm is used to cluster 'm' observations into 'c' clusters. Each cluster has its cluster centre. The observation belongs to the cluster which has the least distance from it. The clustering is crisp, which means that each observation is clustered into one and only one cluster.

FCM is a variation of the hard c-means clustering algorithm. Every observation here has a membership value associated with each of the clusters which is related inversely to the distance of that observation from the centre of the cluster.

### 3.1 The FCM Algorithm

The main aim of FCM algorithm is to minimize the objective function given by [6]:

$$J_m = \sum_{i=1}^{N} \sum_{j=1}^{C} u_{ij}^m \|x_i - c_j\|^2 \quad , \quad 1 \leq m < \infty \tag{2}$$

Where

m denotes any real number greater than 1

$u_{ij}$ *denotes* membership degree of xi in J*th* cluster

$x_i$ is i*th* dimension of the measured data

$c_i$ is the i*th* dimension of the cluster centre.

The formula for updating the cluster centre $c_i$ *is:*

$$c_j = \frac{\sum_{i=1}^{N} u_{ij}^m \cdot x_i}{\sum_{i=1}^{N} u_{ij}^m}$$

(3)

The Partition Matrix values are updated using the formula:

$$u_{ij} = \frac{1}{\sum_{k=1}^{C} \left( \frac{\|x_i - c_j\|}{\|x_i - c_k\|} \right)^{\frac{2}{m-1}}}$$

(4)

The FCM algorithm's iteration stops when the maximum change in the values of Fuzzy c-partition matrix is less than E, where E is a termination criterion with value between 0 and 1.

The steps followed in FCM Algorithm are given below:

1. Initialize the fuzzy partition matrix, $U=[u_{ij}]$ matrix, $U^{(0)}$

2. At k-step: calculate the centre's vectors $C^{(k)} = [c_j]$ with $U^{(k)}$ using equation (3).

3. Update $U^{(k)}, U^{(k+1)}$ using equation (4).

4. If $\| U^{(k+1)} - U^{(k)} \| < \varepsilon$ then STOP; otherwise return to 2.

The documents are clustered based on a representation called "bag of words".

## 4. Bag of Words Representation of a Document

Words are assigned weights according to significance in the document which can be the number of occurrences of that word in that document.

Consider a paragraph:

*"It is a variation of the Hard C-Means clustering algorithm. Each observation here has a membership value associated with each of the clusters which is related inversely to the distance of that observation from the centre of the cluster."*

The Bag of Words representation for the above paragraph will look similar to Table1.

**Table 1: Bag of Words representation for the paragraph**

| Word | Occurrences |
| --- | --- |
| Variation | 1 |
| Hard | 1 |
| Cluster | 2 |
| Observation | 2 |
| ⋮ | ⋮ |
| Centre | 1 |

Apart from single words, we can also use phrases better results. Few examples of these are "engineering student", "computer games", "Tourist attraction", "gold medal" etc. The same Bag of Words technique can be modified to treat the phrases as single words. In our method, we assigned the weights as the number of occurrences of the word in the document normalized by the document length and multiplied by 10,000.

**5. STEPS INVOLVED IN APPLYING FUZZY LOGIC TO TEXT MINING**

The following steps are followed in order to apply fuzzy logic to text mining.

**5.1 Pre-processing of Text**
This phase involves cleaning up the text like removing advertisements etc. Also we have to deal with hyphens, remove tags from documents like HTML, XML etc. Removal of this unwanted text helps improve the efficiency of our algorithm.

**5.2 Feature Generation**
The documents are represented in the "Bag of Words" method. Stop words like "the", "a", "an" etc. are removed. Word Stemming [2] is carried out. Word Stemming refers to representing the word by its root, example the words - driving, drove and driven should be represented by drive. This is important because not all the documents are in the same active or passive voice, or the same tense.

**5.3 Feature Selection**
Here, the features that we will use are selected. This selection of features can either be done before use or based on use. Number of features that will be used for our process is further reduced. We choose only the features which will help us in our process [3][4][5]. In order to cluster separately, the documents relating to sports and politics, presence of names of people in a document will not help much. There can be names in both of those. However, presence of words like "stadium", "ball", "cheerleaders" etc. will help relate the documents to the field of sports. Similarly, words like "election", "democracy", "parliament" will relate the documents to politics category.

In some situations, however, it will be hard to manually point out the features which distinguish documents that should belong to different types. In such situations, we need to find out what features we should use for our clustering. To do this, we can take sample documents which are already known to belong to our required categories. Then the information gained from the analysis of these documents can be applied to point out the differences among documents belonging to different clusters. The word frequency, for various words (normalized by document length), can be calculated for these documents. We choose the words with a significant word frequency variation between different document types.

### 5.4 Clustering

The method used here is FCM clustering algorithm. Each document to be clustered has already been represented as a "bag of words", and from that "bag" only the essential "words" (features) have been kept. The aim now is to cluster the documents into categories (called classes in FCM).Using FCM, the documents having similar frequencies (normalized by document length) of various selected features are clustered together. The number of clusters to be made can be specified at an early stage but this might result in inaccurate results later if this number is not appropriate. This problem is known as Cluster Validity.

Once the FCM execution finishes, either due to accomplishing the pre-specified number of iterations or because of max-change in the Fuzzy Partition Matrix being lesser than the threshold, we obtain the required number of clusters and all the documents have been clustered among them.

### 5.5 Evaluation and Interpretation of Results

From the Fuzzy Partition Matrix, one can find to what degree a given document belongs to each cluster. If the membership value of a document for a given cluster is high, the document can be said to strongly belong to that cluster. However, if all the membership values corresponding to a specific document are almost same (Example .35, .35, .30 among three clusters), it would imply that the document does not quite strongly belong to any of those clusters.

## 6. EXAMPLE

Suppose we are given some documents which we have to cluster into two categories -"sports" and "politics".

We will not present the first two steps that we followed as they are pretty simple. We will start from the third – feature selection. We had no idea about which words (features) we could use to cluster our documents. Hence, we had to find such features.

We took some documents that we knew were related to Sports and Politics respectively. We applied step1 and step 2 on them and then counted the word frequency of various words using eq(1). Then we analysed the differences in these WF values for same words between documents relating to Sports category and the documents relating to Politics category.

The partial result of our analysis is presented in Table 2.

**Table 2: Word Frequencies for Sports and Politics-related documents**

| Word | WF (Sports) | WF (Politics) |
| --- | --- | --- |
| Win | 10.0213 | 8.9012 |
| Stadium | 203.2321 | 7.1214 |
| Democracy | 1.1213 | 140.1213 |
| Ball | 501.6553 | 30.2121 |
| Team | 250.6312 | 80.8452 |
| Candidate | 38.7658 | 40.2313 |
| Campaign | 8.8350 | 9.4213 |

By looking at the Table 2, it was apparent that Words like "Win", "Campaign" are used in similar amounts in documents belonging to both groups. However, words like "Stadium" (203 v/s 7), "Ball" (501 v/s 30), "Team" (250 v/s 80) and "Democracy" (1 v/s 140) could be used to differentiate among documents belonging to Sports and Politics respectively. For the sake of simplicity and an easy example, we chose only 4 words – "Stadium", "Ball", "Team" and "Democracy". On basis of these 4 words, we clustered the given documents into Politics and Sports categories.

Now that we had the documents to be clustered and had selected the features we wanted to use for the clustering process; we could start our clustering process.

Let D= $\{d_1, d_2, d_3…d_n\}$ represent our 'n' documents to be clustered.

Now each of these documents, $d_i$, is defined by the 'm' selected features i.e.

$d_i$= $\{d_{i1}, d_{i2}, d_{i3}..d_{im}\}$

Here, each $d_i$ in the universe D is a m-dimensional vector representing the m selected features which will be normalised in the same way we did for feature selecting by using eq(1). An alternate way to look at this problem is considering each $d_i$ as a point in m-dimensional space representing the features. D, the universe, is a set of points with n elements (documents) in the same sample space.

Now we have to initialise the fuzzy partition matrix. Keeping in mind the constraints that all entries should be in the interval [0, 1] and the columns should add to a total of 1 and considering that we are given 4 documents to cluster:

**Table 3: Fuzzy Partition Matrix**

|           | Doc1 | Doc2 | Doc3 | Doc4 | Doc5 | Doc6 | Doc7 | Doc8 |
|-----------|------|------|------|------|------|------|------|------|
| Cluster 1 | 1    | 1    | 0    | 0    | 1    | 1    | 0    | 0    |
| Cluster 2 | 0    | 0    | 1    | 1    | 0    | 0    | 1    | 1    |

Our initial assumption is that Doc1, Doc2, Doc5 and Doc6 belong to cluster 1 and Doc3, Doc4, Doc7 and Doc8 belong to cluster 2. Now we have to calculate the initial cluster centres. For c=1, that is cluster1, the centre ($V_1$) can be calculated using eq (3). Since, the membership value of Doc3, Doc4, Doc7 and Doc8 in Cluster 1 is 0, we have

$V_{1j} = ((1)x_{1j}+(1)x_{2j}+(1)x_{5j}+(1)x_{6j}) / (1^2+1^2+1^2+1^2) = (x_{1j}+x_{2j}+x_{5j}+x_{6j})/4$

Now, say our word frequencies were like

**Table 4: Word frequencies in given documents (to be clustered)**

|           | Doc1 | Doc2 | Doc3 | Doc4 | Doc5 | Doc6 | Doc7 | Doc8 |
|-----------|------|------|------|------|------|------|------|------|
| Stadium   | 180  | 200  | 5    | 3    | 210  | 7    | 190  | 2    |
| Ball      | 400  | 410  | 20   | 7    | 380  | 10   | 401  | 15   |
| Team      | 200  | 250  | 40   | 35   | 180  | 20   | 170  | 26   |
| Democracy | 1    | 2    | 40   | 38   | 0    | 27   | 5    | 50   |

$V_{11}$ = (180+200+210+7)/4 = 149.25

Similarly, Calculating all $V_{1j}$ we get

$V_1$ = {149.25, 300, 162.5, 7.5} is the centre of cluster 1.

Similarly, $V_2$ = {50, 110.75, 67.75, 33.25} is the centre of cluster 2.

Now calculating the Euclidian distances of each Document vector from both centre clusters:

$D_{11} = ((180-149.25)^2 + (400-300)^2 + (200-162.5)^2 + (1-7.75)^2)^{1/2} = 111.32$

Similarly,

$D_{12}$ = 149.5, $D_{13}$ = 339.5, $D_{14}$ = 352.5, $D_{15}$ = 102.2, $D_{16}$ = 353.4, $D_{17}$ = 109.2, $D_{18}$ = 351.1

Similarly, from the cluster two:

$D_{21} = ((180-50)^2 + (400-110.75)^2 + (200-67.75)^2 + (1-33.25)^2)^{1/2} = 345.10$

$D_{22} = 382.2, D_{23} = 105.1, D_{24} = 118.3, D_{25} = 334.4, D_{26} = 119.6, D_{27} = 339.7, D_{28} = 116.9$

Now that we have distance of each vector from both cluster centres, we will now update the Fuzzy Partition Matrix, by using eq(4).

Note that we have already decided the value of the 'm' in this formula, earlier in the example.

$U_{11} = [1 + (111.32 / 345.10)^2]^{-1} = .90$, $U_{12} = .867$, $U_{23} = .913$, $U_{24} = .898$, $U_{15} = .9146$, $U_{26} = .897$, $U_{17} = .906$, $U_{28} = .901$

The remaining values of the fuzzy partition matrix are updated such that each column adds to 1. We get the updated Fuzzy Partition Matrix as

**Table 5: Updated fuzzy partition matrix after one iteration**

|           | Doc1  | Doc2  | Doc3  | Doc4  | Doc5  | Doc6  | Doc7  | Doc8  |
|-----------|-------|-------|-------|-------|-------|-------|-------|-------|
| Cluster 1 | 0.900 | 0.867 | 0.087 | 0.102 | 0.915 | 0.103 | 0.906 | 0.099 |
| Cluster 2 | 0.100 | 0.133 | 0.913 | 0.898 | 0.085 | 0.897 | 0.094 | 0.901 |

Say our threshold change (stopping condition) was 0.001. Our max change here is 0.133 which is greater than 0.001. Hence we will continue. Now we will repeat the process by again calculating the new cluster centres. But now we will use the updated fuzzy partition matrix values. So, now our centre for cluster 1 will be calculated using

$V_{1j} = (0.900\ x_{1j} + 0.867\ x_{2j} + 0.087\ x_{3j} + 0.102\ x_{4j} + 0.915\ x_{5j} + 0.103\ x_{6j} + 0.906\ x_{7j} + 0.099\ x_{8j}) / (0.900^2 + 0.867^2 + 0.087^2 + 0.102^2 + 0.915^2 + 0.103^2 + 0.906^2 + 0.099^2)$

## 7. INTERPRETING THE RESULTS

Suppose that our final fuzzy partition matrix for 8 documents looks something like this:

**Table 6: Sample Final Fuzzy Partition Matrix**

|           | Doc1  | Doc2  | Doc3  | Doc4  | Doc5  | Doc6  | Doc7  | Doc8  |
|-----------|-------|-------|-------|-------|-------|-------|-------|-------|
| Cluster 1 | 0.890 | 0.804 | 0.149 | 0.155 | 0.865 | 0.159 | 0.810 | 0.168 |
| Cluster 2 | 0.110 | 0.196 | 0.851 | 0.845 | 0.135 | 0.841 | 0.190 | 0.832 |

We see that Doc1, Doc2, Doc5, Doc7 belong to cluster 1 whereas Doc3, Doc4, Doc6, Doc8 belong to cluster 2 on basis of high membership values in those respective clusters.

In the example we took, the documents Doc1, Doc2, Doc5, and Doc7 had higher frequency of words like stadium, ball and team (which are related to sports). Since these documents have been clustered together, one can say that Cluster 1 is Sports. At the same time, Cluster2 will contain documents with higher frequency of words like democracy (which are related to politics). Hence Cluster2 represents Politics.

Here, Doc1 relates to Sports to a large degree (due to its very high membership value). If we set the criteria that membership values greater than 0.85 with respect to a given cluster can be called "strong membership", then Doc1 and Doc5 can be said to "strongly" belong to Cluster1 which is sports. Moreover, we can say that Doc1 relates to Sports more "strongly" than Doc5 does. This interpretation of results in linguistic form [8] is what gives advantage to usage of Fuzzy Logic over Probability models like Bayesian in Text Mining.

## 8. CONCLUSION

In this paper, we showed how one can use fuzzy logic in text mining to cluster documents by taking an example where the documents were clustered into two topics :- "sports" and "politics". The advantage of using fuzzy logic over probability was that in the former, we could calculate the degree to which a given document belonged to either categories- "sports" as well as "politics". This was not possible in the probability model. In other words, instead of simply saying whether the document belonged to "sports" or "politics", we could now tell the degree to which the document characteristics resembled that of a sports-related document as well as a politics-related document. By doing this for all documents in the data-set, we could also compare two documents and tell which one belongs "more" to which topic. Also, because every document will have some membership values in each of the clusters, no useful document will ever excluded from search results [1][5].

## 9. REFERENCES

1. Vishal Gupta, Gurpreet S. Lehal; "A Survey of Text Mining Techniques and Applications"; Journal of Emerging Technologies in Web Intelligence, Vol.1, No.1, August 2009
2. K.Sathiyakumari, V.Preamsudha, G.Manimekalai; "Unsupervised Approach for Document Clustering Using Modified Fuzzy C mean Algorithm"; International Journal of Computer & Organization Trends –Volume 11 Issue3-2011.
3. R. Rajendra Prasath, Sudeshna Sarkar: Unsupervised Feature Generation using Knowledge Repositories for Effective Text Categorization. ECAI 2010: 1101-1102
4. Sumit Goswami, Sudeshna Sarkar, Mayur Rustagi: Stylometric Analysis of Bloggers' Age and Gender. ICWSM 2009
5. Sumit Goswami, Mayank Singh Shishodia; "A fuzzy based approach to stylometric analysis of blogger's age and gender"; HIS 2012: 47-51
6. Ross, T. J. (2010); "Fuzzy Logic with Engineering Applications", Third Edition, John Wiley & Sons, Ltd, Chichester, UK
7. Fuzzy logic vs Proobability (Good Math, Bad Math); http://scientopia.org/blogs /goodmath/ 2011 /02 /02 / fuzzy-logic-vs-probability/, last checked on 28th July 2012.
8. Nogueira, T.M. ; " On The Use of Fuzzy Rules to Text Document Classification "; 2010 10th International Conference on Hybrid Intelligent Systems (HIS),; 23-25 Aug 2010 Atlanta, US.